\documentclass[sigconf]{acmart}
\AtBeginDocument{%
  }

\copyrightyear{2025}
\acmYear{2025}
\setcopyright{cc}
\setcctype{by-nc-nd}
\acmConference[CIKM '25]{Proceedings of the 34th ACM International Conference on Information and Knowledge Management}{November 10--14, 2025}{Seoul, Republic of Korea}
\acmBooktitle{Proceedings of the 34th ACM International Conference on Information and Knowledge Management (CIKM '25), November 10--14, 2025, Seoul, Republic of Korea}
\acmDOI{10.1145/3746252.3761488}
\acmISBN{979-8-4007-2040-6/2025/11}

\usepackage{tabularray}
\usepackage{balance}

\begin{document}

\title[The Temporal Game: A New Perspective on Temporal Relation Extraction]{The Temporal Game: A New Perspective \\ on Temporal Relation Extraction}

\author{Hugo Sousa}
\authornote{Work done before joining Amazon.}
\orcid{0000-0003-3226-9189}
\affiliation{%
  \institution{University of Porto \\ INESC TEC}
  \city{Porto}
  \country{Portugal}
}
\email{hugo.o.sousa@inesctec.pt}

\author{Ricardo Campos}
\orcid{0000-0002-8767-8126}
\affiliation{%
  \institution{University of Beira Interior \\ INESC TEC}
  \city{Porto}
  \country{Portugal}
}
\email{ricardo.campos@inesctec.pt}

\author{Alípio Jorge}
\orcid{0000-0002-5475-1382}
\affiliation{%
  \institution{University of Porto \\ INESC TEC}
  \city{Porto}
  \country{Portugal}
}
\email{alipio.jorge@inesctec.pt}

\begin{abstract}
  In this paper we demo the Temporal Game, a novel approach to temporal relation extraction that casts the task as an interactive game. 
  Instead of directly annotating interval-level relations, our approach decomposes them into point-wise comparisons between the start and end points of temporal entities. 
  At each step, players classify a single point relation, and the system applies temporal closure to infer additional relations and enforce consistency. 
  This point-based strategy naturally supports both interval and instant entities, enabling more fine-grained and flexible annotation than any previous approach.
  The Temporal Game also lays the groundwork for training reinforcement learning agents, by treating temporal annotation as a sequential decision-making task. 
  To showcase this potential, the demo presented in this paper includes a Game mode, in which users annotate texts from the TempEval-3 dataset and receive feedback based on a scoring system, and an Annotation mode, that allows custom documents to be annotated and resulting timeline to be exported. 
  Therefore, this demo serves both as a research tool and an annotation interface.
  The demo is publicly available at \url{https://temporal-game.inesctec.pt}, and the source code is open-sourced to foster further research and community-driven development in temporal reasoning and annotation.
\end{abstract}



\maketitle

\section{Introduction} \label{sec:intro}

The ability to understand how events unfold over time is essential for natural language understanding. 
At the heart of this lies the task of temporal relation extraction, which seeks to determine how events and temporal expressions -- hereafter referred to as \emph{temporal entities} -- are positioned relative to one another in time. 
By identifying these temporal relations, systems can construct coherent timelines, answer time-sensitive questions, and follow the progression of narratives~\cite{Leeuwenberg2019,Campos2021,Su2025}.
Yet, teaching machines to reason about time remains a significant challenge.
This is partially due to the fact that the annotation of temporal relations is a labor-intensive process, often requiring fine-grained judgments that are difficult to automate~\cite{Wang2022,Roccabruna2024}.

Most existing annotation frameworks, such as TimeML~\cite{Pustejovsky2003}, rely on Allen's interval relations~\cite{10.1145/182.358434}, where events are treated as time intervals and annotated with one of 13 possible relations (e.g., before, after, overlaps). 
While expressive, this setup imposes a heavy cognitive burden on the annotators, often leading to low agreement~\cite{Ning2018}. 
Moreover, reasoning with interval-level relations can be computationally expensive and prone to inconsistencies without temporal closure mechanisms~\cite{Vilain1986,Verhagen2005}.

Recent work has shown that decomposing interval relations into point-wise comparisons -- specifically, relations between event start and end points -- can improve inter-annotator agreement~\cite{Ning2018, Alsayyahi2023} and produce effective classification models~\cite{Huang2023}.
By focusing on point-level relations (e.g., whether the start of one event occurs before the end of another), makes the annotation process clearer and less restrictive, enabling the creation of a more precise timeline.
However, this decomposition increases the number of relations that must be annotated.
To mitigate this, temporal closure has been employed to accelerate the annotation of interval relations~\cite{Verhagen2005, UzZaman2011}.
Interestingly, temporal closure is also more precise at the point level, as it reduces ambiguity and limits opportunities for error.
For instance, if interval annotations specify that event A starts both B and C, no interval relation can be inferred between B and C (see~\citet{Leeuwenberg2019} to understand the mapping from interval to point of the start relation).
Yet, in the point-based setting, most point relations can be inferred via transitivity, except for the one between the end points of B and C.
Despite these advantages, tools to support point-based annotation remain limited, and the potential for interactive or learning-driven approaches is still largely unexplored.

In this paper, we present the Temporal Game, a novel approach to temporal relation extraction that frames the task as an interactive game. 
Instead of directly annotating interval relations, users classify the point relations -- which can only take four labels: before, after, equal, and vague -- between the start and end points of the temporal entities. 
At each step, the system computes temporal closure to propagate constraints and infer additional relations. 
This not only reduces annotation effort, but also ensures consistency and enables fine-grained annotations that include both intervals and time instants.

The demo is deployed as a web application and offers two modes: a Game mode, where users play through real examples from the TempEval-3 dataset~\cite{UzZaman2013} with a scoring system and feedback, and an Annotation mode, which supports manual or semi-automatic annotation of custom texts. 
The demo application supports entity editing, dynamic annotation workflows, and downloadable annotations as a JSON file. 
All code is made open source with a permissive license to support further research and development\footnote{\url{https://github.com/hmosousa/temporal_game_demo}}.

Our work serves both as a proof of concept for a point-based annotation pipeline and as a foundation for future reinforcement learning approaches to temporal relation extraction, where an agent could be trained to maximize annotation accuracy or efficiency through interaction with the game. 

\section{Temporal Game} \label{sec:game}
\begin{figure}[!ht]
  \centering
  \includegraphics[width=0.4\textwidth]{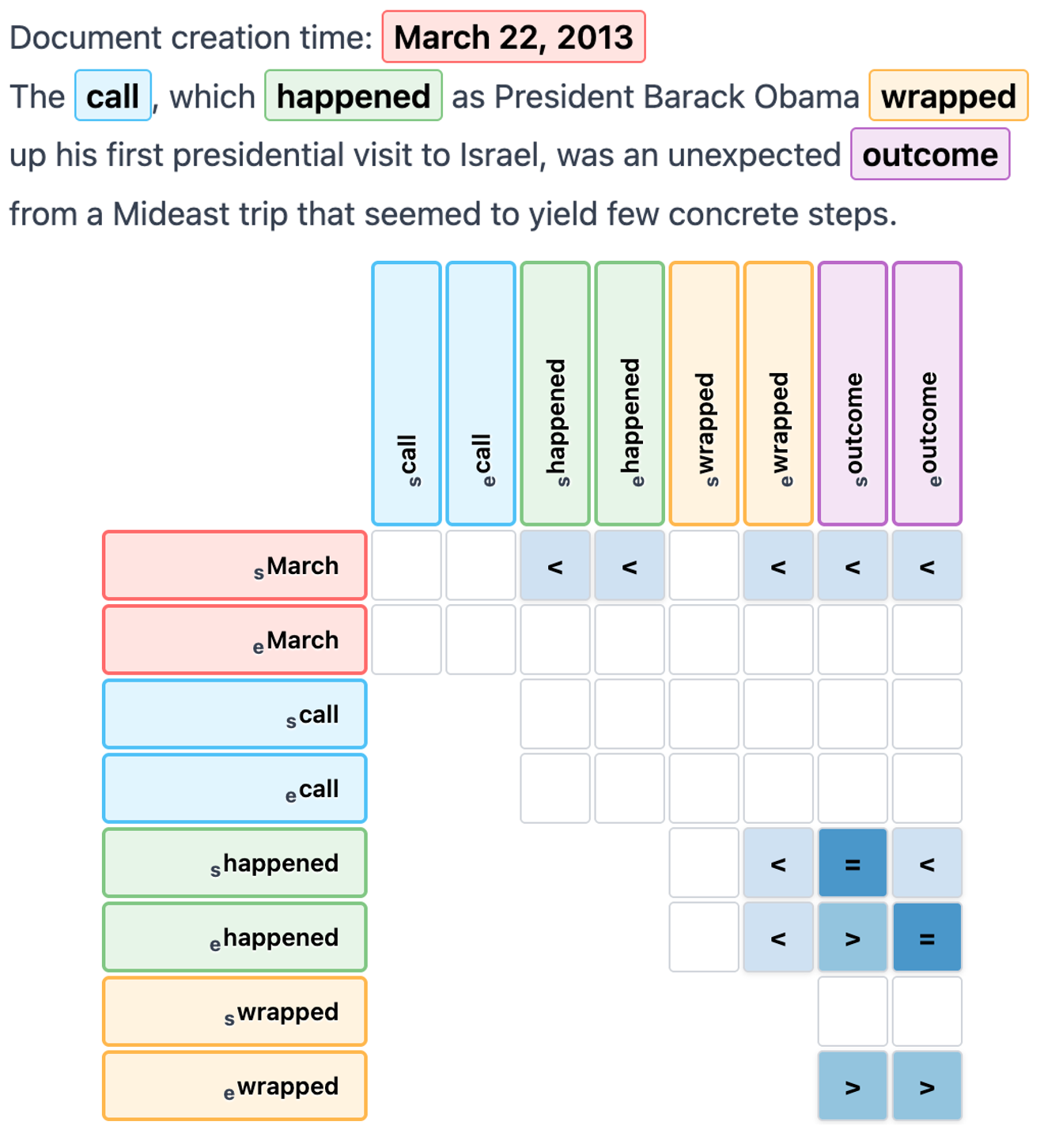}
  \caption{A Temporal Game with five entities being annotated. At the top is the tagged text being annotated, with the temporal entities highlighted. Below is the temporal board, with the entities' endpoints as rows and columns. The white cells represent relations that are yet to be annotated, while the others show the relations already selected by the player.}
  \label{fig:example}
  \Description{Matrix-style board comparing temporal entity end-points.}
\end{figure}

Temporal relation extraction focuses on identifying the temporal relations between a set of entities tagged in a text.
To this end, the Temporal Game was designed with two components: the \emph{tagged text}, which is the text with the highlighted temporal entities for which the temporal relations will be classified, and the \emph{temporal board}, which is a matrix of the entities' endpoints, where each entry has a point relation to be selected.
Figure \ref{fig:example} shows an example of a Temporal Game with five entities being annotated. 

Note that the first entity, ``March 22, 2013'', is the document creation time. 
This is one of the most important entities to annotate, as it is used to ground the other entities in the time the document was created or published.
However, the document creation time is not typically explicit on the text, but in the metadata of the document.
To make this information accessible within the game, we prepend all input texts with the phrase ``Document creation time: <dct>'', where <dct> is replaced by the actual creation time of the document.

Below the text, Figure \ref{fig:example} shows the temporal board, with the entities' endpoints as rows and columns.
The temporal board is initialized as a square matrix of size \( j \times j \), where \( j = 2k + i \), with \( k \) denoting the number of interval entities and \( i \) the number of instant entities.
To reduce redundancy, only the upper triangular portion of the matrix is retained, as the lower triangular part would contain inverse relations of those already represented. 
Additionally, diagonal entries -- corresponding to relations between an entity and itself -- are hidden, since they would always have an equal relation. 
As a result, the first entity is removed from the columns and the last entity is removed from the rows, because all their cells would be hidden. 
For example, in Figure~\ref{fig:example}, the entity ``March 22, 2013'' does not appear in the columns, and ``outcome'' does not appear in the rows.

It is important to note that we use the entities' endpoints -- i.e., the start and end of the entities-- instead of the entities themselves.
This is based on the findings of \citet{Ning2018}, where, through an annotation experiment, the authors found that annotation between the start points of entities achieves higher agreement than annotation between their end points.
Furthermore, it has also been shown that systems trained on the start and end points of entities can yield effective results~\cite{Huang2023}.
This approach is also beneficial because it naturally allows for the integration of time instants -- temporal entities for which the start and end are the same -- rather than relying solely on time intervals.
This is important because using only interval relations, as in TimeML~\cite{Pustejovsky2003,Saur2006}, does not allow for annotation between an interval and an instant.
Therefore, with this approach, we can accommodate a more fine-grained annotation than any previous approach.
On top of that, at the point level there are only four possible relations~\cite{Ning2018}, namely: before (<), after (>), equal (=), and vague (-).
This is a significant reduction from the 13 possible relations at the interval level~\cite{10.1145/182.358434} and therefore makes the annotation more straightforward.

\begin{figure*}[!ht]
  \centering
  \includegraphics[width=\textwidth]{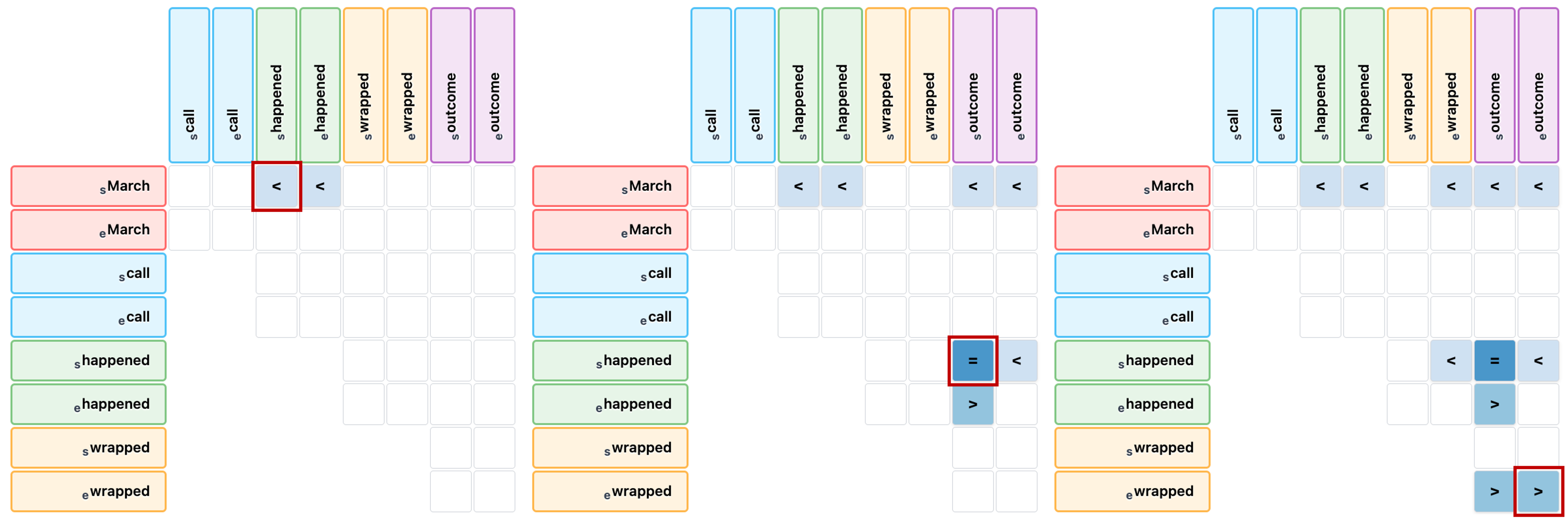}
  \caption{The three boards above show a sequence of annotations made by the user. The cells selected by the user are highlighted with a red square. At each step, the additional relations are inferred through temporal closure.}
  \label{fig:closure}
  \Description{Sequence of annotations made by the user.}
\end{figure*}

On the other hand, decomposing temporal relations into classifications between individual endpoints increases the number of relations that must be annotated, making the annotation process more labor-intensive.
To alleviate this burden, we leverage the transitivity rules of temporal relations to infer additional relations from a smaller set of annotated ones, a process known as temporal closure~\cite{10.1145/182.358434}.
This is illustrated in Figure~\ref{fig:closure}.
In addition to reducing the annotation effort and speeding up the process, temporal closure helps ensure the coherence of the resulting timeline.
Nonetheless, inconsistencies may still arise if the user selects a relation that conflicts with existing ones. Such a contradiction represents one of the game's termination conditions.
The other occurs when the player successfully completes the board without introducing any incoherencies.

\section{Demonstration} \label{sec:demo}

The demo is available at \url{https://temporal-game.inesctec.pt}.
It is a web application that was built using Node.js for the frontend and Flask for the backend.
For the computation of temporal closure we use the \texttt{tieval} Python package~\cite{Sousa2023}.

On the landing page of the demo, the user is presented with two options: Game mode and Annotation mode. In the following sections we describe the two modes in detail.

\subsection{Game Mode} \label{sec:game-mode}
In Game mode, the user is presented with a text that has been annotated with temporal relations. 
These annotations are sourced from the TempEval-3 dataset~\cite{UzZaman2013}, which provides temporal relations at the interval level. 
To prepare the data for our system, we first convert these interval-based annotations to point-level relations using the mapping defined in the SemEval-13 evaluation script\footnote{\url{https://github.com/naushadzaman/tempeval3_toolkit}}. 
We then apply temporal closure at the point level to infer additional relations beyond those explicitly annotated.

To keep the gameplay focused and manageable, we restrict Game mode to single-sentence examples. 
Besides that, the user can choose the number of entities they want to annotate -- ranging from two to five -- which defines the level of the game and controls its difficulty.

To generate these games, we first split each TempEval-3 document into individual sentences and prepend the DCT to each one. 
Then, we count the number of temporal entities in each sentence ($n$) and decide how to use the sentence based on the chosen level $l$. 
If a sentence contains fewer than $l$ entities, it is discarded for that level. 
If it contains exactly $l$ entities, it is used as-is. 
When a sentence contains more than $l$ entities, we generate all possible combinations of $l$ entities from the $n$ available, resulting in $\binom{n}{l}$ different games derived from that single sentence.
After that, we retrieve all generated games and ensure that each includes at least one annotated temporal relation. 
This guarantees that the user is presented with a concise and easy-to-follow text that still contains enough temporal information to evaluate, at the end of the game, whether the produced timeline aligns with the manual annotations.

The statistics of the games generated using this approach are presented in Table~\ref{tab:game-stats}. 
Note that the number of games at level~2 is lower than at other levels, as we require each game to contain at least one point relation. 
Additionally, the vague relation does not appear in any of the games, since no interval relation from TempEval-3 maps to a set of point relations that includes it. 
To address this limitation, one could cross-reference annotations from TimeBank-Dense~\cite{Cassidy2014} to incorporate the vague relation into the games. 
However, we leave this extension for future iterations of the demo.

\begin{table}
    \centering
    \caption{Statistics of the generated games (Section~\ref{sec:game-mode}). Token counts were computed using a whitespace tokenizer.}
    \label{tab:game-stats}
    \begin{tblr}{
        width = \columnwidth,
        colspec = {Xccccc},
        row{3-6} = {r},
        column{1} = {l},
        column{3} = {c},
        column{5} = {c},
        cell{1}{3} = {c=3}{},
        cell{1}{6} = {r=2}{},
        cell{1}{2} = {r=2}{},
        hline{1,7} = {-}{0.08em},
                hline{3} = {-}{0.05em},
            }
                   & \textbf{\#~Tokens } & \textbf{\#~Relations} &            &                         & \textbf{\#~Games } \\
        $l$        &                     & \textbf{\textless{}}  & \textbf{=} & \textbf{\textgreater{}} &                    \\
        \textbf{2} & 410,412             & 19,142                & 2,477      & 26,130                  & 12,928             \\
        \textbf{3} & 1,091,081           & 78,518                & 9,629      & 100,908                 & 30,944             \\
        \textbf{4} & 1,580,590           & 159,636               & 18,584     & 199,637                 & 40,802             \\
        \textbf{5} & 1,630,299           & 213,363               & 24,159     & 265,761                 & 38,932
    \end{tblr}
\end{table}

\begin{figure}[!ht]
  \centering
  \includegraphics[width=0.4\textwidth]{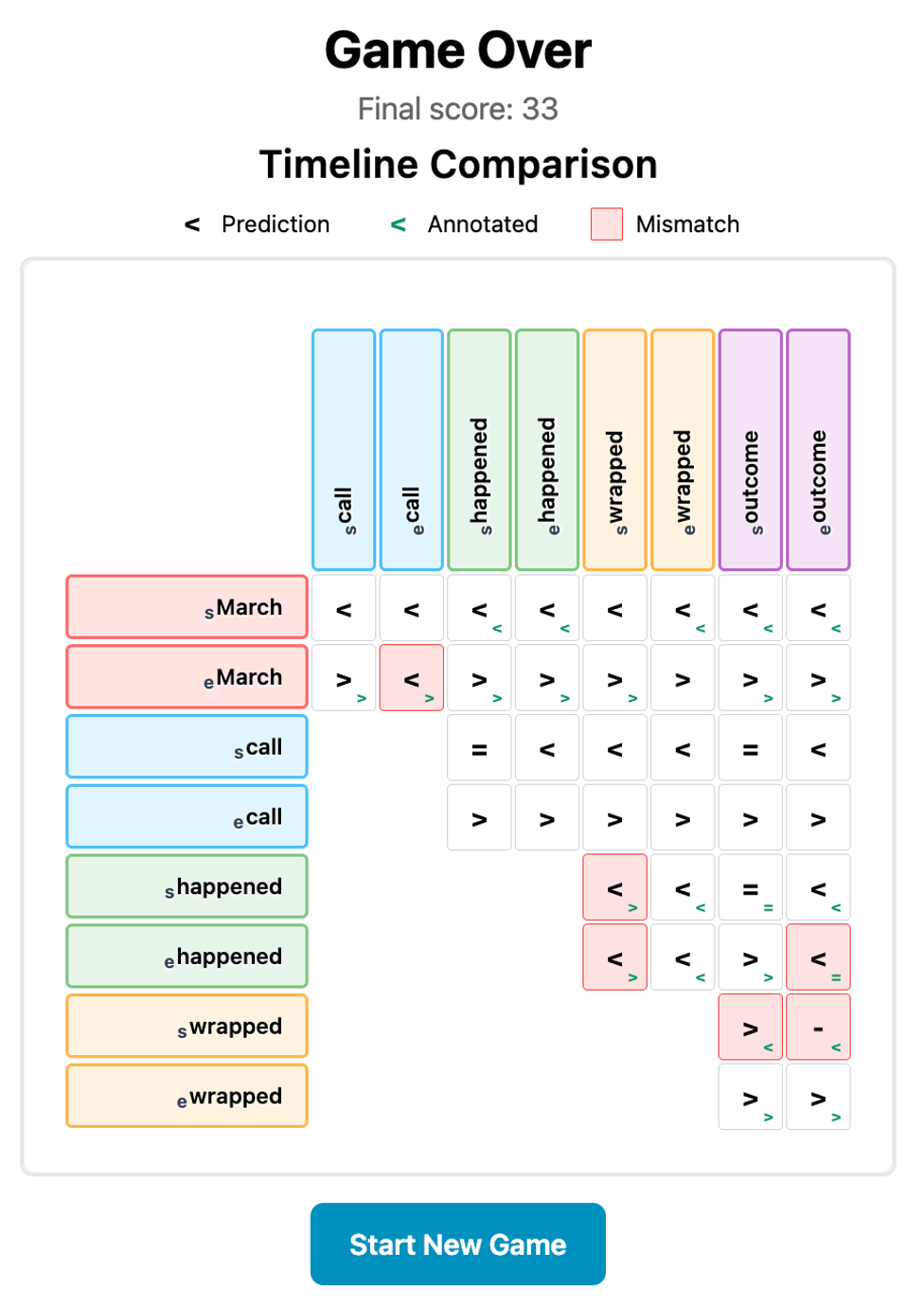}
  \caption{Example of the feedback board shown at the end of a game. Each cell contains the predicted temporal relation (center) and the corresponding gold annotation (bottom right). Discrepancies between the two are highlighted in red.}
  \label{fig:game-over}
  \Description{Example of the feedback board shown at the end of a game.}
\end{figure}

At the end of the game, the user is shown a board that compares their predicted relations with the TempEval-3 annotations (Figure~\ref{fig:game-over}).
A final score is also presented at this stage, calculated as follows:
\begin{itemize}
  \item Step reward:
  \begin{itemize}
  \item +1 if the relation predicted by the user is in accordance with the original annotation
  \item +0.5 if the relation predicted by the user has no value in the original annotation
  \item -1 if the relation predicted by the user is not in accordance with the original annotation
  \end{itemize}
  \item Terminal reward:
  \begin{itemize}
  \item +10 if a coherent timeline is produced
  \item -10 if an incoherent timeline is produced
  \end{itemize}
\end{itemize}

This score system serves to illustrate how one could use the Temporal Game to train a temporal relation extraction agent in a reinforcement learning setting~\cite{Sutton2018}, an approach that, to the best of our knowledge, is yet to be explored.
In this scenario, one could also consider rewarding the agent for the number of relations inferred.
We intend to explore different scoring functions in the future.

At the endgame menu the player is also presented with a button to start a new game which will randomly sample a game from the database and present it to the user.
Alternatively, the user can go back to the landing page and explore the Annotation mode, which is described in the next section.

\subsection{Annotation Mode} \label{sec:annotation-mode}

In this mode, the game experience is the same as in Game mode, but no score is kept.  
Additionally, the system simply flags the annotator if an annotation produces an incoherent timeline, instead of terminating the game.

The main difference in this interface is that the user can upload their own text for annotation.  
This can be done by uploading a simple text file or a JSON file with the following fields: 
\begin{itemize}
  \item \texttt{dct}: the document creation time.
  \item \texttt{text}: the text to be annotated.
  \item \texttt{entities}: a list of dictionaries with the start and end character offsets of the entity in the text.
\end{itemize}

Of these fields, only \texttt{text} is mandatory.
If the \texttt{dct} field is provided, the text will be prepended with ``Document creation time: <dct>'' as in the Game mode. 
Otherwise, the text will be annotated as is.
If the \texttt{entities} field is not provided, the system provides an \texttt{Annotate Entities} button that will automatically detect entities using the baseline event extractor from \texttt{tieval}~\cite{Sousa2023}, and temporal expressions using the \texttt{TEI2GO} model~\cite{Sousa2023_tei2go}.  
However, this functionality is limited to English text.

Alternatively, the user can manually highlight entities in the text.  
This is achieved by clicking and dragging the mouse over the desired text span.  
This action adds the selected entity to the board.  
Users can remove an entity by hovering over it and clicking the delete button.  
Entities can also have their type changed between \textit{interval} and \textit{instant}.  
By default, all entities are marked as intervals.  
To change this, the user can click on the entity and select the new type from a dropdown menu.  
This will update the board by replacing the start and end entries of the entity with a single entry, prepended with an \textit{i}, indicating an instant.

Since an annotator might want to annotate a large document with many entities, the board can quickly become cluttered and difficult to navigate.  
To address this, the annotator can toggle the \textit{dynamic mode}.  
In this mode, the interface presents a temporal board for only one endpoint pair at a time.  
The pair of entities to be displayed can be selected either randomly or in a guided manner.  
In the random setting, the system selects a pair of entities that are missing an annotation at random.  
In the guided setting, we use confidence scores from a temporal relation classification model to select the pair~\cite{TemporalClassifierModel}.  
Pairs for which the model is most confident are shown first.  
The rationale for this is that the model is most confident about the relations that are easier to annotate.
This should, in principle, lead to a more efficient annotation process.

At the end of the annotation, the user can download the annotations in a JSON file by clicking the export button.

\section{Conclusions \& Future Work} \label{sec:conclusions}

In this demonstration paper, we introduce the concept of the Temporal Game, a novel approach to temporal relation extraction where the task is framed as a game in which the temporal relations between entity endpoints are classified one at a time.
This not only presents a new way to train a temporal relation extraction system in a reinforcement learning setting, but also offers a novel approach to annotating temporal relations.
In the future, we plan to conduct annotation experiments to assess whether this approach achieves higher annotation agreement than the interval-based approach.
Furthermore, we also plan to train a reinforcement learning model to play the game and evaluate its effectiveness.

\begin{acks}
  This work is funded by national funds through FCT - Fundação para a Ciência e a Tecnologia, I.P., under the support UID/50014/2023, project \url{https://doi.org/10.54499/2022.14691.BD}, and project \url{https://doi.org/10.54499/2022.09312.PTDC}.
\end{acks}

\newpage

\section*{GenAI Usage Disclosure}
Generative AI was used during the development of the Temporal Game demo to generate part of the frontend code.
It was also used as a search tool to find relevant work to this research.
Besides that, we also used generative models to find typos and errors in the final manuscript.


\bibliographystyle{ACM-Reference-Format}
\balance
\bibliography{references}

\end{document}